\documentclass[letterpaper, 10 pt, conference]{ieeeconf}  

\IEEEoverridecommandlockouts                              

\usepackage{graphicx} 
\usepackage{multirow}
\usepackage{multicol}

\usepackage{amssymb}
\usepackage{pifont}
\newcommand{\cmark}{\ding{51}}%
\newcommand{\xmark}{\ding{55}}%

\usepackage{amsmath}
\usepackage{gensymb} 
\usepackage{subcaption}

\usepackage{booktabs}
\usepackage{siunitx}
\usepackage{comment}

\usepackage{caption}
\usepackage{afterpage}

\title{\LARGE \bf
Event-Only Drone Trajectory Forecasting with RPM-Modulated Kalman Filtering
}

\author{Hari Prasanth S.M.$^{1*}$, Pejman Habibiroudkenar$^{1*}$, Eerik Alamikkotervo $^{1}$ , Dimitrios Bouzoulas $^{1}$ and Risto Ojala  $^{1}$ 
\thanks{*Equal contribution}
\thanks{This work was supported by Aalto University}
\thanks{$^{1}$Department of Energy and Mechanical Engineering,
        Aalto University, 02150 Espoo, Finland
        {\tt\small firstname.lastname@aalto.fi}}%
}

\begin{document}

\maketitle
\thispagestyle{empty}
\pagestyle{empty}

\begin{abstract}

Event cameras provide high-temporal-resolution visual sensing that is well suited for observing fast-moving aerial objects; however, their use for drone trajectory prediction remains limited. This work introduces an event-only drone forecasting method that exploits propeller-induced motion cues. Propeller rotational speed are extracted directly from raw event data and fused within an RPM-aware Kalman filtering framework. Evaluations on the FRED dataset show that the proposed method outperforms learning-based approaches and vanilla kalman filter in terms of average distance error and final distance error at 0.4s and 0.8s forecasting horizons. The results demonstrate robust and accurate short- and medium-horizon trajectory forecasting without reliance on RGB imagery or training data.

\end{abstract}

\section{INTRODUCTION} \label{sec:introduction}

Trajectory forecasting of Unmanned Aerial Vehicles (UAVs) is an essential requirement for several use cases such as airspace monitoring, collision avoidance, autonomous aviation, and anti-drone technologies. While detection and tracking are useful, predicting the future trajectory of an aerial object enables early decision-making and threat analysis.
This is particularly challenging in real-world scenarios, where drones can accelerate rapidly and drastically change direction. Another layer of difficulty is added if the drone is non-cooperative, and there is no access to its control signals, or on-board measurements and forecasting can only rely on remote detections. 

While non-cooperative drone detection and forecasting have been researched extensively, most of the current methods focus on RGB cameras \cite{isaac2021unmanned, ma2024research,becker2021generating}.
Unfortunately, all these methods can experience motion blur, limited frame rates, and latency in sensor measurements, which can compromise the accuracy of state observations, especially in aggressive maneuvers or high-speed flights.
As trajectory forecasting requires accurate and reliable past state observations, these sensing limitations affect the performance of any approach that uses RGB camera.
\begin{figure}[t]
    \centering
    
    \includegraphics[width=\linewidth]{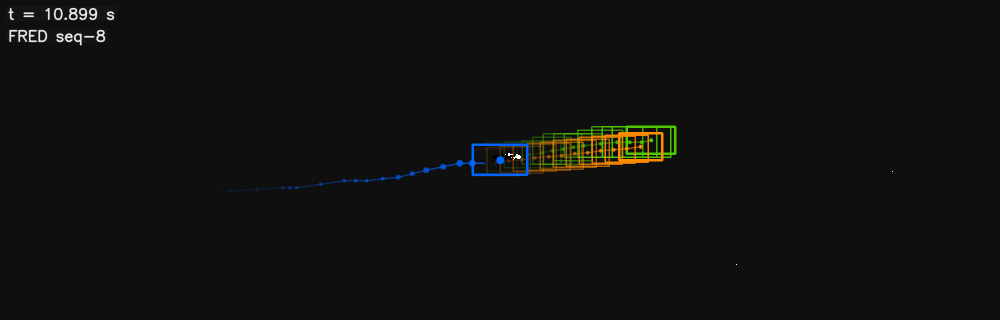}
    \vspace{-0.95em}
    
    \includegraphics[width=\linewidth]{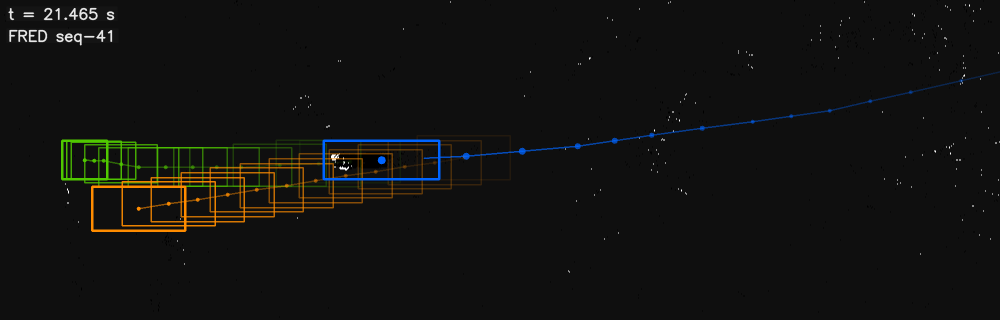}
    \vspace{-0.95em}
    
    \includegraphics[width=\linewidth]{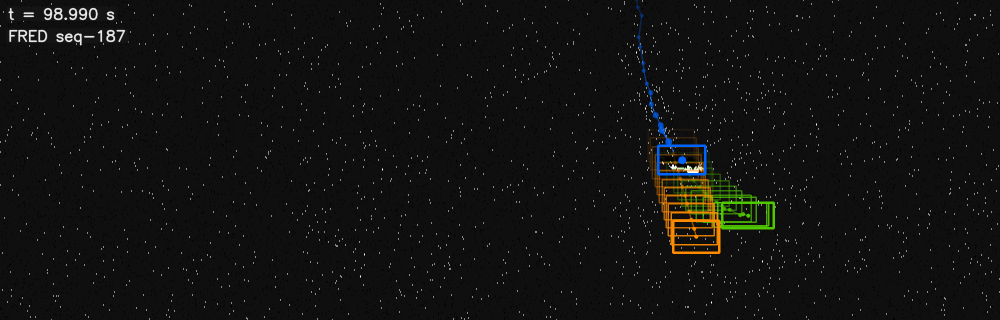}
    
    \caption{\textbf{Overview of the proposed trajectory forecasting up to 0.4\,s into the future.}
Blue denotes the current drone position and past trajectory. Future ground-truth positions are shown in green, and predictions in orange.}
    
    \label{fig:overview}
\end{figure}

Event cameras have been proposed as an alternative sensor solution to address this high-speed vision challenge.
Unlike conventional cameras, event cameras operate asynchronously at the per-pixel level.
Each pixel reacts to a relative change in the intensity level of the observed scene at the order of microseconds~\cite{gallego_event-based_2022}.
These capabilities are useful to avoid motion blur when observing fast motions, and in recent years, there has been growing interest in event cameras for event-based perception in robotics and drone applications~\cite{mueggler2017event, gallego_event-based_2022}. Event cameras have employed for drone presence detection~\cite{eldeborg2024drone,stewart2022virtual}, localization~\cite{magrini2024neuromorphic} and tracking~\cite{iaboni2022event, mitrokhin2018event} but their potential for trajectory prediction remains mostly unexplored. 

Remote trajectory forecasting typically relies on kinematic modelling~\cite{peng2018unified} or machine learning based solutions~\cite{wang2024enhanced,becker2021generating,magrini_fred_2025}. While machine learning methods can learn complex prediction mechanisms from data, they need large amounts of training data and the learned prediction mechnanism transfer poorly to drone types and flight patterns that are not included in the training data. 

Deep Learning (DL) based event camera trajectory forecasting has been presented in \cite{magrini_fred_2025}, but the performance gains over non event-camera based configurations where quite low. This paper presents a method to extract critical drone properties from event camera data, such as propeller speed with classic algorithms to improve the state estimation of the drone. The future states can then be predicted from the current state using kinematic motion modelling. Here Kalman Filtering is employed for state estimation and the detected propeller rotation speed is used for dynamic real-time estimation of the process noise. The proposed method is evaluated with the currently most extensive event camera based drone dataset, FRED \cite{magrini_fred_2025}, where we show improved forecasting accuracy for both 0.4s and 0.8s prediction horizons over the DL based baseline and vanilla Kalman Filter based approach. Examples of forecasted trajectories are presented in Figure~\ref{fig:overview}. The contributions can be summerized as: 
\begin{itemize}
    \item We introduce a real-time Kalman Filter based trajectory forecasting method with propeller rotation speed modulated process noise matrix for improved performance. 
    \item We present state of the art results for event only trajectory forecasting on the FRED dataset \cite{magrini_fred_2025} outperforming DL based baseline and vanilla Kalman Filter.
\end{itemize}

\section{RELATED WORK}\label{sec:related_work}


Trajectory Forecasting is the task of predicting the future poses using only information available up to the current moment. Drones are typically highly agile with erractic movements, making the forecasting task extremely challenging. A short overview of the most common drone trajectory prediction techniques is provided in sections \ref{sec:on-board-prediciton} and \ref{sec:remote-prediction}. For a more detailed review, we recommend \cite{shukla2024trajectory}. In section \ref{subsec:event_based}, event cameras and their applications to drone detection and forecasting are introduced. A more detailed review can be found in \cite{magrini_drone_2025}. 

\subsection{Drone Trajectory Forecasting with On-board Sensors}\label{sec:on-board-prediciton}
If the drone's onboard sensor information is available and drone properties are known, the future trajectory can be estimated with dynamic modelling. It estimates the forces and torques acting on the drone to derive its future state. The most important forces for a drone are the aerodynamic drag, gravity, and propeller thrust. Dynamic modelling of drones has been explored in \cite{kamel2017dynamic,chovancova2014mathematical}. In this paper, we focus on non-cooperative drone trajectory forecasting where dynamic modelling is typically not feasible. 

\subsection{Remote Drone Trajectory Forecasting}\label{sec:remote-prediction}
For a remote observer of a non-cooperative drone kinematic modelling can be used instead of dynamic modelling.  Kinematic modelling assumes constant velocity or acceleration over the forecasting interval to integrate the future state. 
For drones, the constant velocity or acceleration assumptions typically only hold for short time intervals. Kinematic modelling as a remote observer is used in \cite{peng2018unified}. 

Alternatively, the trajectory prediction task can be learned from data using classical machine learning or DL methods. An example of a classical Gaussian Process-based trajectory forecasting method is presented in \cite{xie2022efficient}. Most DL methods use sequential data from the past trajectory to predict the future trajectory, and a variety of architectures like transformers~\cite{wang2024enhanced}, Recurrent Neural Networks (RNN) \cite{becker2021generating}, and Long-Short Term Memory (LSTM) \cite{magrini_fred_2025} have been proposed. While DL models can learn advanced prediction patterns from data, they require high computation and large amounts of high-quality training data, and generalisation outside the training domain is typically poor. 

\subsection{Event Cameras and Drone Trajectory Forecasting}\label{subsec:event_based}
Traditional cameras capture absolute intensity at every single pixel at each frame, which makes them vulnerable to challenging lighting conditions and fast-moving objects. On the other hand, event cameras capture changes in brightness for each pixel individually with microsecond-level accuracy and only generate a sample for pixels where the change in intensity is over a given threshold. A continuous stream of data is produced where each sample, referenced as an event, includes the spatial pixel coordinate, microsecond resolution timestamp, and polarity of the intensity change: if the intensity decreased or increased. This working principle enables significantly higher dynamic range and allows capturing high-speed motion without suffering from motion blur. A more detailed survey on event cameras is provided in \cite{gallego_event-based_2022}.

The properties of event cameras are highly suitable for detecting drones, and event camera-based drone perception has become an increasingly popular topic of research \cite{magrini_drone_2025}. Promising results have been presented for drone presence detection \cite{eldeborg2024drone,stewart2022virtual}, drone localization \cite{magrini2024neuromorphic}, drone tracking \cite{iaboni2022event}, and drone propeller blade analysis \cite{zhao2022high,sanket2021evpropnet}. However, event-camera-based drone trajectory prediction has been little researched. To our knowledge, event-camera-based trajectory prediction with real event camera data has been presented only in \cite{magrini_fred_2025}, where a Convolutional Neural Network (CNN) - Transformer hybrid approach was used. Past event camera frames were cropped to the drone and encoded with a CNN, and these encodings were then fed to the transformer together with the past trajectory states to predict the future states. However, this approach provided quite modest improvements over models that didn't have access to the event camera data. We believe that the prediction accuracy could be significantly improved by extracting information like propeller rotation speed from the event camera data and utilizing it for improved state estimation and motion forecasting. 

\section{METHODOLOGY}\label{sec:methods}
\begin{figure}[!t]
    \centering
    \includegraphics[width=\columnwidth]{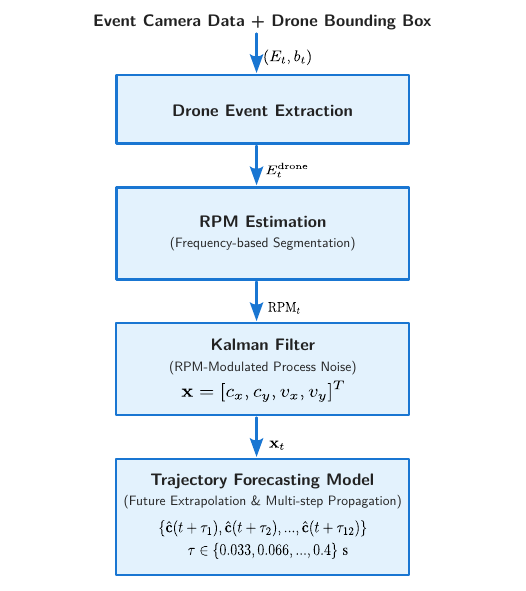}
    \caption{\textbf{Overview of the proposed event-based trajectory forecasting framework.}}
    \label{fig:pipeline}
\end{figure}

The proposed framework uses event camera data and drone bounding box as inputs. Although the bounding box can be obtained from any detector models, we use ground-truth annotations to evaluate trajectory forecasting independently of detection performance. Ground-truth annotations from past trajectory are also used in \cite{magrini_fred_2025}. Our approach leverages the high temporal resolution of event camera to extract the propeller rotation speed,  which is then used to modulate the Kalman Filter process noise. An overview of the complete framework is shown in Figure \ref{fig:pipeline}.


\subsection{RPM Estimation}\label{subsec:rpm_estimation}
Propeller rotation frequency encodes critical information about thrust and maneuverability. We estimate revolutions per minute (RPM) from the temporal periodicity of propeller events using a period histogram approach. 
To construct a spatial frequency map $F(x,y)$, we count the number of events for each pixel events within the drone bounding box.
For accurate RPM estimation the airframe pixels must then be separated from the propeller pixels. 
By selecting a percentile-based threshold $\tau_p$, we differentiate high-frequency propeller pixels from the low-frequency airframe pixels. In experiments, 70th percentile yields best results. This percentile-based threshold enables our system to adapt to changes in the event rate caused by different lighting conditions and flight speeds. Examples of the frequency map and final segmented propellers are provided in Figure \ref{fig:freq_heatmap}. 


\afterpage{%
\begin{figure*}[!t]
    \centering
    \includegraphics[width=\textwidth]{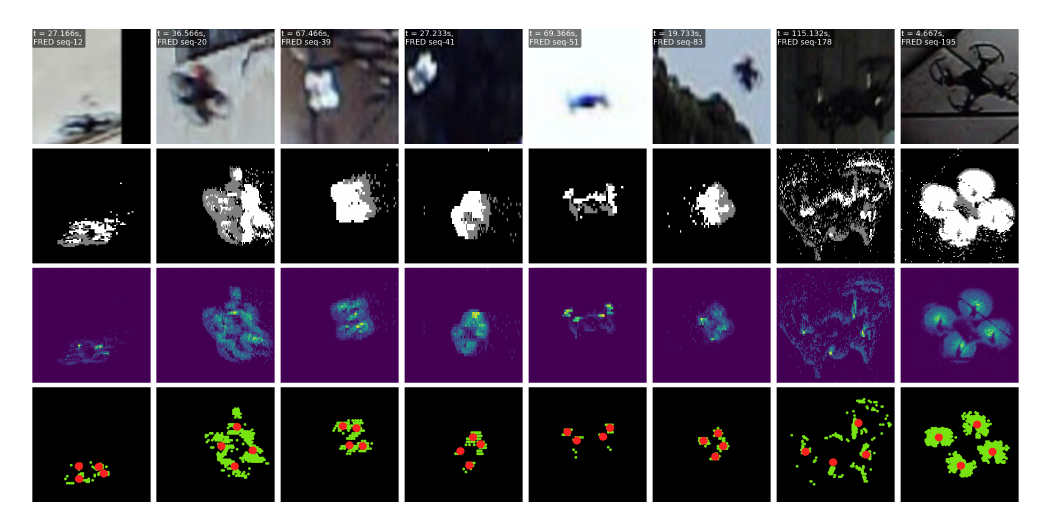}
    \caption{\textbf{Multi-scene visualization of rotor localization from event data.} Each column corresponds to a different scene. From top to bottom: RGB detection, corresponding event frame, frequency heatmap of drone events, and filtered propeller-only events, where red markers indicate the estimated rotor centers. }
    \label{fig:freq_heatmap}
\end{figure*}
}

Each filtered propeller pixel $(x, y)$ maintains timestamps of its last ON and OFF events: $t_{ON}(x, y)$ and $t_{OFF}(x, y)$. If a pixel is turned ON and it was previously turned OFF, the time between the two events $\Delta t$ is recorded. The periods of all pixels inside the bounding box are binned into a histogram $H(b)$ with bin width of $\Delta b = 0.1ms$ and a range of $[0, 25.6ms]$ resulting in total of 256 bins. To account the RPM changes over time, histogram entries older than $100ms$ are removed via First In First Out queue. The most prominent period $\hat{T}$ is identified as:

\begin{equation*}
    \hat{T} = (b^* + 0.5) \cdot \Delta b, \quad b^* = \arg\max_b H(b)
\end{equation*}

The dominant frequency $f = 1/\hat{T}$ is converted to RPM as:

\begin{equation*}
    \text{RPM} = f \cdot \frac{60}{N_b}
\end{equation*}

where $N_b$ is the number of blades per propeller. The values for bin width and number of bins are experimentally chosen and with the chosen values RPM can be estimated in the range 2300 to 300,000 RPM.



\subsection{Trajectory Forecasting Model}\label{subsec:traj_forecast_model}
We formulate trajectory prediction as a real-time Kalman Filtering problem that estimates the current state from noisy measurements and predicts future positions using kinematic motion model. The process noise matrix is modulated based on the detected propeller rotation speed as high propeller rotation speeds correlate with eratic movements.

\subsubsection{Motion model}
The drone state is estimated in the image space and it includes four state variables:
center position $(c_x, c_y)$ and velocity $(v_x, v_y)$. 
We follow the discrete-time constant velocity for the state transition model \cite{montanez2023application, lizzio2023comparison}. The motion model assumes that $v_x$ and $v_y$ remain constant over prediction interval and $c_x$ and $c_y$ are updated as follows: 

\begin{align*}
    c_{x,k} &= c_{x,k-1} + v_{x,k-1} \Delta t \\
    c_{y,k} &= c_{y,k-1} + v_{y,k-1} \Delta t 
\end{align*}

The process noise covariance matrix is initialized as a diagonal matrix and scaled by a coefficient that depends on the detected propeller rotation speed. Higher propeller RPM is associated with more aggressive maneuvers and rapid accelerations, so the filter increases motion uncertainty and gives more weight to the measurement update.
Conversely, at low RPM the drone motion is assumed to be smoother and closer to hovering and more weight can be given to the motion model. 

The process noise covariance matrix is initialized as a diagonal matrix and
scaled by a coefficient that depends on the detected propeller rotation speed.
Higher propeller RPM is associated with more aggressive maneuvers and rapid
accelerations, so the filter increases motion uncertainty and gives more weight
to the measurement update. Conversely, at low RPM the drone motion is assumed
to be smoother and closer to hovering, allowing greater reliance on the motion model.

The velocity-related scaling factor is defined as
\begin{equation*}
\alpha_v = \max\!\left(0.5,\; 1 + 2r + \max(0,\dot r)\right),
\end{equation*}

where $r \in [0,1]$ denotes the normalized RPM level and reflects the available thrust and overall maneuvering capability of the drone. 
The term $\dot r \in [-1,1]$ captures the rate of change of RPM, where positive values indicate increasing thrust and a higher likelihood of rapid motion changes. 
Consequently, the process noise increases both with higher steady thrust and during rapid throttle increases, allowing the estimator to better handle dynamic motion while remaining conservative during stable flight.

\subsubsection{Measurement model}

All state variables ($c_x, c_y, v_x,v_y$) are measured directly. Center position $(c_x,c_y)$ is recovered from the ground truth bounding box center and velocity components $(v_x,v_y)$ are calculated from the difference between ground truth bounding box position at current and previous timestep.  

\subsubsection{Forecasting}
The state at current time $t$ is recovered from the Kalman Filter with RPM modulated process noise. The forecasted state at $t+ \tau$ is predicted using the Kalman Filter motion model with timestep $\tau$. Naturally the measurement update is not performed for the forecasted state. Predictions are made for 0.4s and 0.8s horizons to follow the benchmark format of the FRED dataset \cite{magrini_fred_2025}. In addition to the final position at the end of the interval, all intermediate poses are forecasted with 33ms timesteps for evaluation of the full forecasted trajectory. 

\subsection{FRED Dataset}\label{subsec:fred_dataset}

To evaluate the proposed method, we use the FRED Dataset~\cite{magrini_fred_2025}.
FRED is a benchmark designed for drone detection, tracking, and trajectory forecasting, combining synchronized RGB video and event camera data at HD resolution (1280$\times$720).
The dataset contains over 7 hours of annotated recordings per modality (more than 14 hours in total), making it the largest RGB-event dataset specifically focused on aerial drone scenarios.
The data are split into two sections: one for training and one for testing. Since our method is not learning-based, we utilize only the test split.

FRED includes five different drone models, ranging from ultra-light FPV mini-drones to commercial drones, thereby covering a wide spectrum of object sizes, flight speeds, maneuverability, and hovering behaviors.
It provides dense per-frame annotations with bounding boxes and consistent track identities, enabling not only object detection but also multi-object tracking and trajectory forecasting.
The recordings span a diverse set of challenging real-world scenarios, including rain, low-light and nighttime conditions, indoor environments, long-range flights, multiple simultaneously visible drones, and visual distractors such as insects.
Furthermore, FRED offers both canonical and domain-shifted challenging splits, which are specifically designed to evaluate model robustness and generalization under adverse conditions.

\begin{figure*}[!t]
    \centering
    \includegraphics[width=\textwidth]{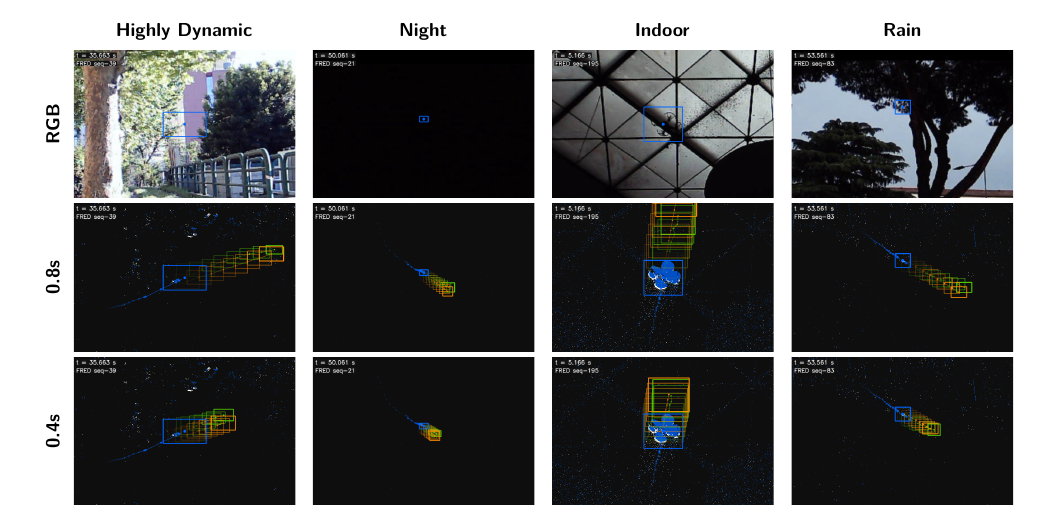}
    \caption{\textbf{Qualitative trajectory forecasting results under challenging conditions.} The current ground-truth position and past trajectory are shown in blue; future ground-truth positions are shown in green; predicted future positions are shown in orange. Results are illustrated across highly dynamic motion, night-time, indoor, and rain scenarios.}
    \label{fig:qualitative}
\end{figure*}

\subsection{Evaluation}\label{subsec:evaluation_metrices}
The propose method is evaluated against classic and deep learning based forecasting methods. All models are evaluated with the FRED test set and deep learning based models are trained only with the train set that is separate from test samples. 
LSTM, Transformer, and the CNN + Transformer models are deep learning based models presented in \cite{magrini_fred_2025}. LSTM and Transformer use only the past ground-truth bounding box information as input, while CNN + Transformer models additionally use past RGB and/or event frames encoded with CNN. 
As the classical baselines we implemented Linear extrapolation and vanilla Kalman Filter.
Linear extrapolation estimates the $(v_x, v_y)$ velocities directly as the average of four previous poses without any filtering and then forecasts the future poses with the same constant velocity motion model that is used in our method. 
Vanilla Kalman Filter is implemented identically to our method, except the process noise matrix is fixed and not modulated based on the rotor rotation speed. 

Following the FRED benchmark protocol, forecasting performance is evaluated using two standard trajectory prediction metrics: Average Displacement Error (ADE) and Final Displacement Error (FDE).
ADE measures the mean Euclidean distance between the predicted trajectory and the ground-truth trajectory over all predicted time steps, while FDE computes the Euclidean distance between the predicted final position and the corresponding ground-truth final position. Both metrics are reported in pixel space, where lower values indicate better forecasting accuracy.  Since our method is primarily focused on trajectory forecasting rather than bounding box regression, predictions are evaluated using the center point of the object rather than full bounding box coordinates. This choice allows for a fair and direct comparison of trajectory prediction accuracy across different methods, independent of object scale or aspect ratio variations.

\section{Results}\label{sec:experiments}

ADE and FDE for the 0.4s and 0.8s forecasting intervals for the evaluated methods are presented in Table~\ref{table_main_results}. Our proposed method achieves the highest performance in all metrics. All classical methods overperform the deep learning based methods and the performance gap is larger for the 0.4s horizon. The highest-performing deep learning based model is the CNN + Transformer utilizing both RGB and event camera data. 

Kalman Filter achieves 0.1px improvement for 0.4s horizon and 2.2px for the 0.8s horizon over the linear extrapolation baseline. Modulating the Kalman Filter process noise based on the propeller rotation speed yields additional improvements of 1.6px for 0.4s horizon and 4.2px for 0.8s horizon. 

\begin{figure}[h]
    \centering
    \includegraphics[width=\linewidth]{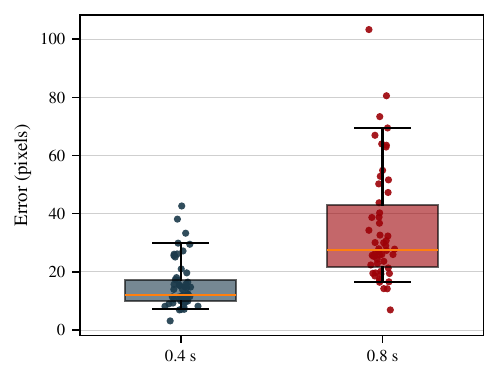}
    \vspace{0.1em}
    \includegraphics[width=\linewidth]{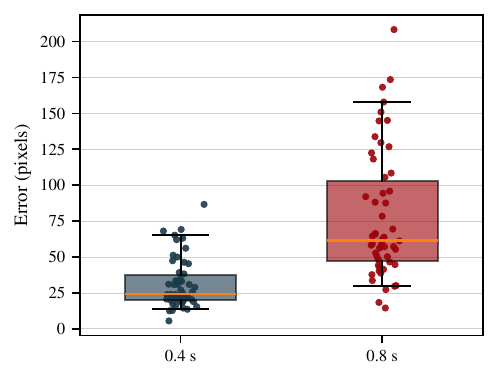}
    \caption{Distribution of forecasting errors at 0.4\,s and 0.8\,s horizons over 54 test cases.
    Top: ADE, Bottom: FDE.
    The box represents the interquartile range (25th–75th percentile), whiskers denote the 5th–95th percentile, and the orange colored line indicates the median.}
    \label{fig:forecast_boxplots_vertical}
\end{figure}

To further evaluate the error distribution of our method the distribution of ADE and FDE across all test sequences for both prediction horizons are presented in 
Figure~\ref{fig:forecast_boxplots_vertical}. 
The boxplots illustrate not only the mean forecasting performance but also the variability.

\begin{table}[!h]
\centering
\caption{Trajectory forecasting results on the FRED dataset. ADE and FDE are reported in pixels.}
\label{table_main_results}

\setlength{\tabcolsep}{4pt}
\begin{tabular}{lcc|cc|cc}
\toprule

\multirow{2}{*}{\textbf{Method}} 
& \multicolumn{2}{c|}{\textbf{Input}} 
& \multicolumn{2}{c|}{\textbf{ADE (px)} $\downarrow$} 
& \multicolumn{2}{c}{\textbf{FDE (px)} $\downarrow$} \\

\cmidrule(lr){2-7}

& RGB & Event 
& 0.4s & 0.8s 
& 0.4s & 0.8s \\

\midrule

LSTM                & \xmark & \xmark 
& 171.9 & 353.3 
& 53.04 & 96.44 \\

Transformer         & \xmark & \xmark  
& 124.8 & 292.2 
& 46.74 & 87.68 \\

CNN + Transformer   & \cmark & \xmark    
& 124.0 & 280.9 
& 45.92 & 83.86 \\

CNN + Transformer   & \xmark & \cmark 
& 151.7 & 337.0 
& 58.06 & 100.6 \\

CNN + Transformer   & \cmark & \cmark 
& 121.3 & 281.5 
& 45.23 & 85.78 \\

\midrule

Linear Extrapolation  & \xmark & \xmark 
& 16.45 & 37.93 
& 32.90 & 83.32 \\

Kalman Filter    & \xmark & \xmark 
& 16.40 & 37.08 
& 32.80 & 81.54 \\

\textbf{Kalman + RPM} & \xmark & \cmark 
& \textbf{15.35} & \textbf{34.85} 
& \textbf{31.16} & \textbf{77.76} \\

\bottomrule
\end{tabular}
\end{table}

\section{Discussion}\label{sec:disscussion}

Our proposed method achieved the highest forecasting peformance at both 0.4s and 0.8s horizon on the extensive FRED drone dataset.  
Our approach utilizes only event camera data as input and does not rely on learning-based models, requiring little to no parameter tuning. 
In our experiments (Table \ref{table_main_results}) even a simple linear extrapolation yields better forecasting performance than the best performing deep learning model. This suggests that the deep learning models may suffer from overfitting to the training data, limiting its generalization ability. Furthermore, the results demonstrate that incorporating a Kalman Filter for state estimation while keeping the same motion model leads to consistent improvements in both ADE and FDE compared to linear extrapolation. These gains are further enhanced when the Kalman Filter is combined with RPM modulated process noise. These results underline the significance of accurate state estimation in forecasting. The accuracy of the current state strongly affects the accuracy of the forecasted state.  

The trajectory forecasting performance in different scenarios, namely highly dynamic, night, indoor and rain is illustrated in Figure~\ref{fig:qualitative}.
The proposed method achieves good forecasting performance in all of these scenarios. Additionally the figure highlights the advantage of using event camera based perception. The drone is clearly visible in the event camera frames while very difficult to detect in most RGB frames.    

\section{Conclusion}\label{sec:conclusion}
This study presented a real-time event-only approach for drone trajectory prediction that detects the propeller rotation speed from the event camera data and leverages that to dynamically modulate the Kalman Filter process noise matrix. The proposed method operates without reliance on RGB imagery and is well suited for high-speed and visually challenging conditions. Experimental results on the FRED dataset demonstrate that our approach significantly outperforms learning-based methods in both short- and medium-horizon. These findings highlight the potential of classic event camera based methods for accurate UAV motion forecasting in real-world scenarios.

Future work will focus on integrating individual propeller RPM estimation, instead of single RPM estimate for whole drone to enable more precise trajectory forecasting, particularly during rapid and erratic movements. In addition, the orientation of the drone could be estimated based on the detected propeller for improved state estimation and forecasting. These values can also be utilized for various drone applications other than forecasting, such as behavioral analysis.  
Additionally, we plan to evaluate the performance of our method in real-world flight experiments with drones operating in unconstrained environments.

\section*{ACKNOWLEDGMENT}
The authors would like to thank Junction for organising the hackathon and Sensofusion for introducing the usecase challenge that motivated the development of this article. 

\bibliographystyle{plain} 
\bibliography{reference_manual}

\end{document}